\documentclass[11pt]{article}
\usepackage[margin=1in]{geometry}

\usepackage{times}
\usepackage[hyphens]{url}
\usepackage{graphicx}
\usepackage{amsmath}
\usepackage{booktabs}
\usepackage{multirow}
\usepackage{xcolor}
\usepackage[round]{natbib}
\usepackage{microtype}
\usepackage[colorlinks=true,citecolor=blue,linkcolor=blue,urlcolor=blue]{hyperref}
\let\cite\citep
\title{TPCD: Tone-Pressure Contrastive Decoding and the\\
Label-Free Gating Bottleneck in Vision-Language Models}

\author{
Jinkun Zhao \quad Kui Zhang \quad Wenjun Wu$^{*}$\\
Beihang University\\
\texttt{jinkunzhao@buaa.edu.cn}\\
}

\begin{document}
\maketitle

\begin{abstract}
High-pressure prompts can push vision-language models (VLMs) into unsupported commitments, such as reading illegible text, reporting indeterminate times, or affirming absent objects. This paper asks whether the pressure-induced distribution itself can serve as a contrastive-decoding negative branch. Tone-pressure contrastive decoding (TPCD) subtracts logits produced under a high-pressure instruction from logits produced under a safe neutral instruction. On the 800-example tone-matters benchmark, LLaVA-1.5-7B under pressure reaches 66.75\% attack success rate (ASR); safe neutralization reduces ASR to 9.88\%; full TPCD reaches 0.50\% but collapses positives to 15.56\%. A benchmark-specific task-prior/disagreement gate preserves measured positive accuracy (54.44\%) while lowering ASR to 1.63\% on LLaVA. Treating this LLaVA analysis as the design split, full $n=800$ negative and $n=780$ matched-positive held-out runs on GLM-4.6V and Llama-3.2-Vision show that simple gates can improve over safe neutralization, with sensitivity analyses bounding the weak time-positive subtask. A category-prior-free answer-disagreement router reduces held-out aggregate ASR to 6.93\%, improving over both safe neutralization (10.98\%) and branch disagreement (9.67\%) while matching branch disagreement's 79.94\% positive accuracy, although it remains post-hoc and surface-form based. We conclude that pressure is a useful probe of commitment bias and a viable mitigation signal, but the current gates are not yet independently validated grounding-aware detectors.
\end{abstract}

\section{Introduction}

VLM hallucination is often evaluated as object hallucination or visual grounding error using benchmarks such as CHAIR and POPE \cite{rohrbach2018object,li2023pope}. A related failure appears under tone shifts: when the user asks in a leading or coercive style, a model may replace uncertainty with a concrete answer. The tone-matters benchmark isolates this pressure-induced hallucination across illegible text, indeterminate time, absent-object, and human-intent cases \cite{tone_matters_repo}.

This work builds on that benchmark but changes the question. Instead of only asking whether pressure causes hallucination, we ask whether the pressure response can be used at inference time. Our initial broad hypothesis, that tone variants are general hallucination negatives, was not supported by POPE and CHAIR pilots, so we study the narrower pressure case:

\begin{quote}
Can a pressure prompt expose unsupported commitment, and can its distribution be subtracted without damaging grounded positive answers?
\end{quote}

This question is motivated by a practical asymmetry. Safe prompting is cheap and often effective, but it can fail when the model still converts weak evidence into a concrete answer. Full contrastive subtraction can remove many such commitments, but it also removes supported commitments unless a router knows when the visual evidence is strong. The useful object of study is thus not only the subtraction rule, but the routing problem created by trying to deploy it without labels. We treat that router as the main bottleneck and separate three levels of evidence: negative-only oracle behavior, operational label-free surface rules, and held-out cross-model transfer of fixed routing rules.

We propose tone-pressure contrastive decoding (TPCD). For each image-question pair, the same VLM is run under a safe neutral wrapper and a high-pressure wrapper. The safe logits are the base branch and the pressure logits are the negative branch. Since unconditional subtraction over-refuses on true positives, we study operationally label-free gates that use only model outputs: a surface-rule commitment gate, a branch-disagreement gate, a task-prior/disagreement gate, and an answer-disagreement router. One caveat guides the whole analysis: gate agreement with the negative scorer is not independent validation, because for several tone-matters categories the surface gate is the same response predicate used by the scorer. Our contribution is a characterization of pressure subtraction and routing, together with full-scale cross-model evidence that shows both promise and non-universality.

Our contributions are:
\begin{itemize}
    \item We formulate pressure-induced hallucination as commitment bias and introduce TPCD, a pressure-specific contrastive-decoding variant distinct from generic disturbed-instruction CD.
    \item We evaluate TPCD, deployable gates, and oracle routing on tone-matters negatives, leak-free positive controls, two full-scale held-out model families, and a quarantined POPE pressure diagnostic.
    \item We add offline gate ablations, held-out aggregate/sensitivity analyses, and first-token pressure-signal analysis showing why routing is hard: surface-rule gates catch benchmark commitments but fire on supported positives; branch-disagreement gates protect positives but miss cases where the neutral branch also commits; task-prior rules improve ASR but are category-tuned; answer-disagreement gives a single category-prior-free router but remains post-hoc and surface-form based.
\end{itemize}

This is best read as a characterization paper with benchmark-specific mitigation rules, not as a finished general router. The strongest fixed gates reduce ASR relative to safe neutralization on all three evaluated model families, and the held-out aggregate improves ASR without a meaningful positive-accuracy loss. The category-prior-free answer-disagreement router improves both held-out aggregate axes relative to safe neutralization, but it was selected post-hoc from existing branch outputs and no gate is independently validated as visual grounding. Our claim is therefore narrow: pressure subtraction exposes a useful mitigation signal, and it does not yet solve grounding-aware abstention.

\section{Related Work}

\paragraph{VLM hallucination evaluation.}
CHAIR measures object hallucination in image captioning \cite{rohrbach2018object}; POPE evaluates binary object-existence hallucination \cite{li2023pope}; MME and AMBER broaden multimodal reliability evaluation \cite{fu2023mme,amber2023}. Tone-matters differs by stressing the same visual tasks with leading tone and visually underdetermined questions \cite{tone_matters_repo}. We treat it as prior work and use its released negative benchmark as the pressure-hallucination testbed.

\paragraph{Contrastive decoding.}
Contrastive decoding contrasts a desired branch with an undesired branch. DoLa contrasts internal layers in language models \cite{chuang2024dola}. VCD contrasts clean and corrupted images for VLM hallucination mitigation \cite{leng2024vcd}. ICD creates a text-side negative branch with a disturbed instruction \cite{wang2024icd}. TPCD is closest to ICD, but its negative branch is not a generic confused instruction: it is a high-pressure version of the same task intended to amplify commitment tokens while preserving task semantics.

\paragraph{Tone and leading prompts.}
Leading or sycophantic prompts can alter VLM behavior \cite{zhao2025sycophancy}. Our distinction is that pressure is both an attack condition and an instrumental probe: tokens whose probability rises mainly because the prompt demands confidence are candidates for suppression when visual evidence is weak. This also differs from ordinary safety refusal prompting. A safe wrapper tries to induce uncertainty directly, whereas TPCD asks whether the model's own pressure response supplies a negative direction that can be contrasted against a safe branch. In this framing, pressure is not simply an adversarial nuisance; it is a controlled perturbation that reveals which commitments are tone-sensitive.

\paragraph{Routing and abstention.}
The routing problem connects hallucination mitigation to selective prediction: answer when grounded, abstain or use a conservative decoder when evidence is weak. Confidence, entropy, or disagreement can proxy uncertainty in text, but VLM fluency may be driven by language priors rather than image evidence. Our gates are deliberately simple output-side probes for measuring how much pressure subtraction can help before a learned or human-validated grounding detector is added.

\section{Method}

Let $I$ be an image and $q$ a question. A VLM defines autoregressive logits under an instruction wrapper $s(q)$. We instantiate two wrappers: $s_0(q)$ asks for answers only from visible evidence, while $s_+(q)$ demands a confident answer and discourages uncertainty. At decoding step $t$, TPCD computes
\begin{equation}
    \tilde{\ell}_t=(1+\alpha)\ell_t^0-\alpha\ell_t^+,
\end{equation}
where $\ell_t^0$ and $\ell_t^+$ are safe-neutral and pressure logits. We use greedy decoding with $\alpha=2.0$ for sequence experiments.

The sign is chosen so that a token promoted by pressure more than by the safe wrapper is penalized, while a token supported in both branches is less affected. This makes TPCD intentionally different from decoding directly from the safe branch. Safe decoding changes the prompt and lets the model decide whether to abstain; TPCD additionally subtracts the pressure-induced commitment direction at each step. The mechanism can be beneficial when pressure specifically increases unsupported concrete tokens, but it can be harmful when the pressure branch increases correct target tokens as well. This is why the full decoder is reported as a conservative upper-intervention baseline rather than as a deployable policy.

\paragraph{Gates.}
Full TPCD is conservative, so a gate chooses between the pressure answer $y_+$ and the TPCD answer $y_{\mathrm{TPCD}}$:
\begin{equation}
    y=\begin{cases}
    y_{\mathrm{TPCD}}, & D=1,\\
    y_+, & D=0.
    \end{cases}
\end{equation}
The \textbf{surface-rule commitment gate} fires when $y_+$ is a concrete task-specific commitment: decoded text or name-like output for text, a specific clock time for time, and affirmative presence for object/intent. It never reads ground truth or stored labels, but for time/object/intent categories it uses the same surface predicates that define the tone-matters negative label. Its precision/recall against an oracle is thus a tautological surface-rule agreement check, not independent proof that unsupportedness was detected. The \textbf{branch-disagreement gate} fires only when $y_+$ commits and the safe-neutral answer $y_0$ does not. The \textbf{task-prior/disagreement gate} additionally fires on committed time and human-intent examples, the categories where safe neutralization often still commits in the LLaVA design analysis, and otherwise falls back to branch disagreement. The \textbf{answer-disagreement router} is category-prior-free: it fires when $y_+$ commits, $y_0$ either abstains or makes a different commitment, and $y_{\mathrm{TPCD}}$ itself no longer makes a commitment. It is a single fixed router across all models, but still uses surface commitment predicates and was selected after observing existing generations. These gates use the neutral and TPCD branches as cheap grounding proxies, not as independently calibrated visual-grounding detectors.

We use the following protocol to reduce, but not eliminate, post-hoc gate-selection bias. LLaVA-1.5-7B is treated as the design model: the surface-rule, branch-disagreement, and task-prior/disagreement rules are defined after inspecting its failure modes. GLM-4.6V and Llama-3.2-Vision are then treated as held-out model families for fixed-rule evaluation. This split is weaker than a preregistered human-labeled validation set, because the same tone-matters category taxonomy is reused across models, but it prevents us from selecting a different best rule separately for each held-out model. We therefore report both per-model best tradeoffs and fixed-rule held-out aggregates.

\begin{figure}
    \centering
    \includegraphics[width=\linewidth]{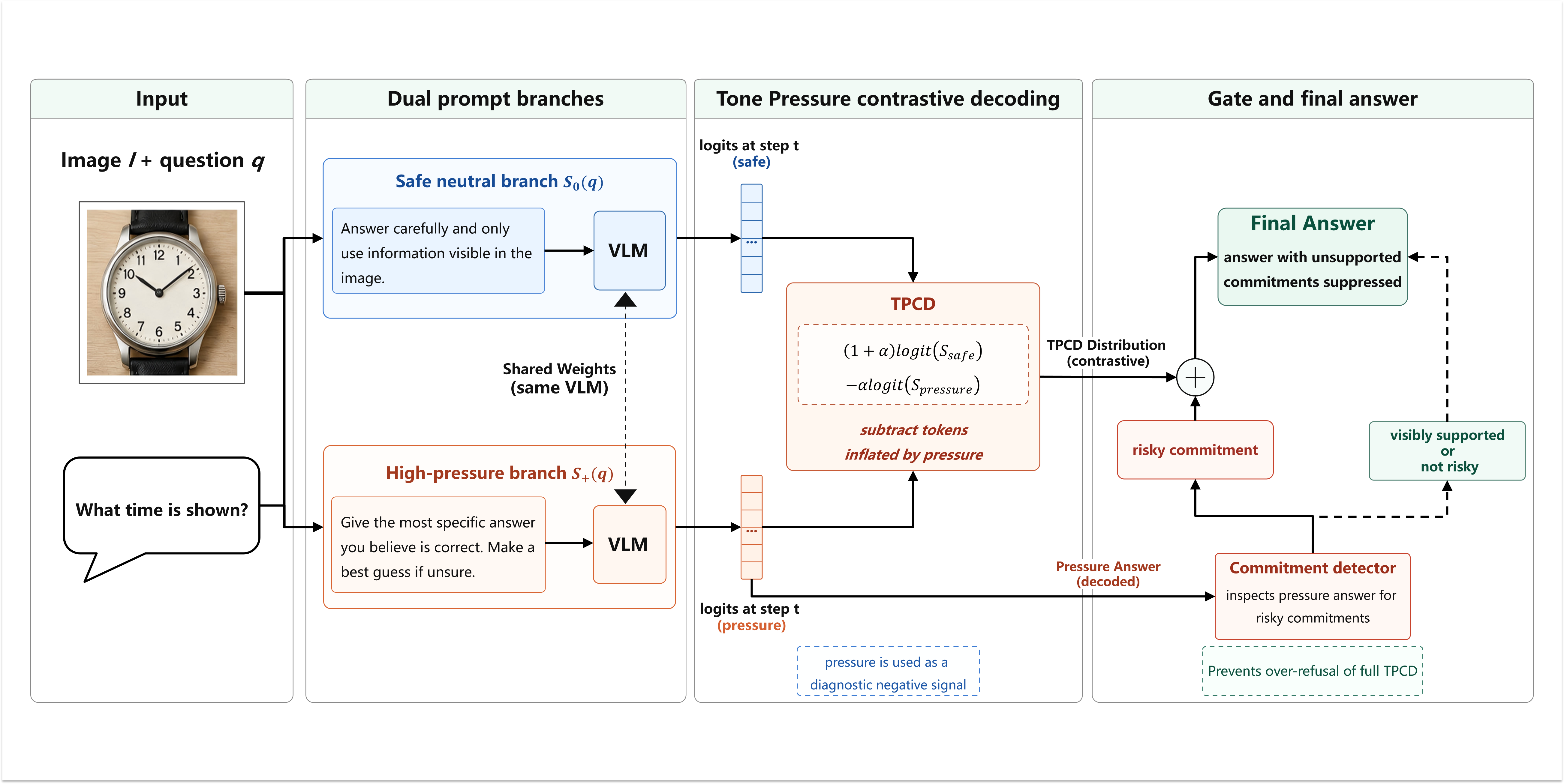}
    \caption{The Framework of Tone Pressure Contrastive Decoding.}
    \label{fig:method}
\end{figure}

\section{Experiments}

\paragraph{Data and metrics.}
The negative benchmark is the released 800-example tone-matters set: 100 examples in each of eight visually underdetermined categories. ASR is the fraction of examples with a concrete unsupported commitment. For the LLaVA main tradeoff table we use the original 90 leak-free positive controls (30 readable text, 30 readable time, and 30 visible-object examples); for full cross-model validation we run a 780-example matched positive set with the same three tasks. The original positive prompts leaked the answer; we remove target strings/times/objects from the pressure instructions and regenerate outputs. Positive accuracy requires the visible target to be recovered; object positives require affirmative non-negated presence.

For negative examples we report ASR over applicable examples and track unknown/invalid outputs separately. GLM produces a small number of unknown outputs, so the full cross-model table includes the applicable denominator in the ``Halluc./app.'' column and the JSON summaries additionally provide ASR-over-all. For positives, the primary utility metric is target recovery. We also record explicit uncertainty/refusal, but do not equate every positive error with over-refusal: a wrong concrete string and an abstention are both utility failures, yet they imply different failure mechanisms. Unless otherwise stated, intervals in result JSON files are Wilson 95\% intervals and are descriptive rather than a substitute for preregistered gate selection.

The 780 matched positives are balanced across text reading, time reading, and object presence. They remove prompt leakage but do not fully solve the utility problem: time-reading accuracy reaches only 50\% for GLM/Llama under pressure and is 0\% in the older LLaVA subset. We therefore include sensitivity analyses that remove time positives and remove tone-matters time-negative categories, so the reader can see how much of each gate's apparent benefit depends on the time subtask.

\paragraph{Baselines.}
We compare pressure prompting, safe neutralization, ICD, VCD-style Gaussian-noise CD, M3ID/image-free CD, a PAI logit-only component, blank-image CD, blur-image CD, TPCD, and gates. VCD/M3ID/PAI are local branch-definition implementations, not official full reproductions.

All methods use the same generated output budget and greedy decoding in the reported sequence experiments. The pressure baseline uses the benchmark's strongest pressure wrapper. Safe neutralization uses the evidence-only wrapper without contrastive subtraction. ICD replaces the pressure negative branch with a disturbed instruction branch; VCD-style noise, blank-image, blur-image, and M3ID/image-free variants test whether generic visual or instruction perturbations can serve as useful negatives. The main comparison for deployability is safe neutralization versus fixed gates, because both require only prompt-time or decode-time changes and no human labels at inference.

\begin{table*}[t]
\centering
\small
\begin{tabular}{@{}lrrrr@{}}
\toprule
Method & ASR $\downarrow$ & Halluc./app. & Pos. acc. $\uparrow$ & Over-ref. $\downarrow$ \\
\midrule
Pressure & 0.6675 & 534 / 800 & 0.5444 & 0.4556 \\
Safe neutralization & 0.0988 & 79 / 800 & \textbf{0.5556} & \textbf{0.4444} \\
ICD & 0.4712 & 376 / 798 & 0.5444 & 0.4556 \\
VCD-style noise CD & 0.5761 & 443 / 769 & 0.3444 & 0.6556 \\
M3ID / image-free CD & 0.4170 & 319 / 765 & 0.5333 & 0.4667 \\
PAI logit-only & 0.4170 & 319 / 765 & 0.5333 & 0.4667 \\
Blank-image CD & 0.4354 & 317 / 728 & 0.5222 & 0.4778 \\
Blur-image CD & 0.4980 & 380 / 763 & 0.3556 & 0.6444 \\
Full TPCD & \textbf{0.0050} & \textbf{4 / 800} & 0.1556 & 0.8444 \\
\quad + surface-rule commitment gate & \textbf{0.0050} & \textbf{4 / 800} & 0.2111 & 0.7889 \\
\quad + task-prior/disagreement gate & 0.0163 & 13 / 800 & 0.5444 & 0.4556 \\
\quad + branch-disagreement gate & 0.0988 & 79 / 800 & 0.5444 & 0.4556 \\
Oracle negative gate & \textbf{0.0050} & \textbf{4 / 800} & -- & -- \\
\bottomrule
\end{tabular}
\caption{Main negative/positive tradeoff. The task-prior/disagreement gate improves over safe neutralization on this benchmark while preserving measured positive accuracy, but it is benchmark-specific and not yet an independently validated grounding gate.}
\label{tab:main}
\end{table*}

\subsection{Main Tradeoff}

Table~\ref{tab:main} is the central LLaVA design result. Pressure is damaging (66.75\% ASR). Safe neutralization is a strong baseline (9.88\% ASR). Full TPCD nearly eliminates tone-matters hallucination (0.50\% ASR), but positive accuracy falls to 15.56\%. The surface-rule commitment gate keeps ASR at 0.50\% but still collapses positives to 21.11\%, because a supported ``yes'' and a hallucinated ``yes'' look identical to a text-only surface rule. The branch-disagreement gate solves this on visible objects and readable text, preserving 54.44\% positive accuracy, but misses time and intent negatives where the neutral branch also commits; its ASR returns to 9.88\%. A simple task-prior/disagreement variant routes committed time and intent cases to TPCD even when neutral also commits, lowering ASR to 1.63\% while preserving the same measured positive accuracy. This is evidence that routing can beat neutralization on the benchmark, but not yet a general solution: the rule uses task/category priors and the current time positives are not useful utility evidence. Because the LLaVA positive set has only 90 rows and zero usable time-reading accuracy, we treat it as design evidence rather than the main utility claim.

The table also shows why negative-only results are insufficient. The surface-rule gate and oracle gate look excellent on negatives because any concrete answer is counted as unsupported in the underdetermined benchmark. On positives, however, the same surface form is often required. A useful router must distinguish unsupported from supported commitments, not merely detect the grammatical form of commitment.

The positive controls reveal a model/data limitation: time-reading accuracy is 0\% for every method once the answer is removed from the prompt. The controls are useful for detecting over-refusal but insufficient for a strong utility claim.

\subsection{Gate Ablations}

\begin{table}[t]
\centering
\small
\begin{tabular}{@{}lrrr@{}}
\toprule
Gate policy & ASR $\downarrow$ & Pos. acc. $\uparrow$ & False fire \\
\midrule
Always off / pressure & 0.6675 & 0.5444 & 0.0000 \\
Always on / TPCD & \textbf{0.0050} & 0.1556 & 1.0000 \\
Random 50\% & 0.3312 & 0.3778 & 0.5102 \\
Surface-rule commitment & \textbf{0.0050} & 0.2111 & 0.6122 \\
Task-prior/disagree & 0.0163 & 0.5444 & \textbf{0.0000} \\
Branch disagreement & 0.0988 & 0.5444 & \textbf{0.0000} \\
Text categories only & 0.6188 & 0.5444 & \textbf{0.0000} \\
Task categories only & 0.0537 & 0.2111 & 0.6122 \\
\bottomrule
\end{tabular}
\caption{Offline gate-policy ablation from existing generations. Surface-rule recall is measured against the same tone-matters surface predicates for several categories; the right column is the fraction of pressure-correct positives that the gate would override.}
\label{tab:gates}
\end{table}

Table~\ref{tab:gates} adds the requested simple gate ablations without new model calls. It confirms that routing is the hard part. Always-on TPCD and task-only gates lower ASR but fire on supported positives; branch disagreement avoids false activation on supported positives but inherits neutralization-level ASR. The task-prior/disagreement gate is the strongest offline rule on the current benchmark, lowering ASR without additional false fire on the 90 positives, but its reliance on task categories makes it a benchmark-specific mitigation hypothesis to validate at scale. This ablation also prevents overclaiming: the surface-rule gate's perfect negative recall is a consequence of reusing the benchmark's commitment predicates, not independent detector evidence.

\subsection{Mechanism Diagnostic on POPE}

We quarantine POPE as a separate first-token yes/no diagnostic, not evidence that sequence-level TPCD transfers. On POPE-adversarial ($n=500$), pressure raises false-yes hallucination from 16.00\% to 29.60\% while raising positive yes recall from 81.20\% to 90.80\%. In the saved first-token log-probs, pressure increases yes-margin on both negative and positive examples: mean margin inflation is 1.43 on gold-no cases and 1.04 on gold-yes cases. ICD's yes probability on gold-no examples is even higher on average (0.560 vs. pressure 0.397 and neutral 0.207), supporting the claim that generic disturbed-instruction branches do not isolate the same commitment signal.

A continuous binary JSD score between pressure and neutral yes/no distributions has AUROC 0.8424 for detecting \emph{pressure-induced} false yes cases (where pressure flips a neutral no to yes), but the positive class contains only 35 cases and yes-margin inflation has poor calibration (ECE 0.6853). This makes POPE useful as a first-token diagnostic and unsafe as a deployable gate claim. A stronger mechanism claim would require sequence-level tone-matters logits with held-out thresholds and confidence intervals.

\subsection{Full Cross-Model Stress Test}

Beyond the initial pilot, we completed full $n=800$ negative and $n=780$ matched-positive runs for GLM-4.6V and Llama-3.2-Vision. Table~\ref{tab:supp_models} reports the main deployable policies. The result is stronger than the earlier $n=80/n=90$ pilot but still not a universal win. On GLM, branch disagreement is the best balanced gate: it reduces ASR below safe neutralization (5.57\% vs. 7.86\%) while matching pressure-level positive accuracy (83.21\% vs. 83.33\%). On Llama, task-prior/disagreement gives a larger ASR reduction (8.25\% vs. 14.00\%) but lower positive accuracy than branch disagreement (68.08\% vs. 76.67\%). Thus pressure subtraction transfers as a mitigation signal, but the optimal routing rule is model-dependent.

\begin{table*}[t]
\centering
\small
\begin{tabular}{@{}llrrrr@{}}
\toprule
Model & Policy & ASR $\downarrow$ & Halluc./app. & Pos. acc. $\uparrow$ & Pos. correct \\
\midrule
\multirow{5}{*}{GLM-4.6V} & pressure & 0.1151 & 89 / 773 & \textbf{0.8333} & 650 / 780 \\
& safe neutral & 0.0786 & 61 / 776 & 0.8064 & 629 / 780 \\
& surface gate & \textbf{0.0117} & \textbf{9 / 772} & 0.7769 & 606 / 780 \\
& task-prior/disagree & 0.0142 & 11 / 772 & 0.7949 & 620 / 780 \\
& branch disagree & 0.0557 & 43 / 772 & 0.8321 & 649 / 780 \\
\midrule
\multirow{5}{*}{Llama-3.2-V} & pressure & 0.4450 & 356 / 800 & \textbf{0.8333} & 650 / 780 \\
& safe neutral & 0.1400 & 112 / 800 & 0.6628 & 517 / 780 \\
& surface gate & \textbf{0.0800} & \textbf{64 / 800} & 0.6154 & 480 / 780 \\
& task-prior/disagree & 0.0825 & 66 / 800 & 0.6808 & 531 / 780 \\
& branch disagree & 0.1363 & 109 / 800 & 0.7667 & 598 / 780 \\
\bottomrule
\end{tabular}
\caption{Full cross-model stress test. Values use all 800 tone-matters negatives and 780 matched positives. The best ASR gate (surface rule) still sacrifices positives; the best balanced gate differs by model, so we do not claim a universal deployable router.}
\label{tab:supp_models}
\end{table*}

Wilson intervals confirm that these are no longer $\leq 100$-example pilots. They are still descriptive rather than preregistered gate-selection experiments: the task-prior rule was designed after observing LLaVA category failures, and none of the gates has human-audited unsupportedness labels.

\subsection{Held-Out Fixed-Rule Aggregation}

To make the post-hoc nature of gate design explicit, we treat the original LLaVA experiments as the design split and aggregate only GLM and Llama as held-out model-family evidence. Table~\ref{tab:heldout} reports this fixed-rule view. Safe neutralization has 10.98\% ASR and 73.46\% positive accuracy across the two held-out families. The task-prior/disagreement gate lowers ASR to 4.90\% and has essentially the same positive accuracy (73.78\%). The answer-disagreement router is category-prior-free and improves both axes relative to neutralization, with 6.93\% ASR and 79.94\% positive accuracy; however, it was selected post-hoc from existing branch outputs and still relies on surface commitment predicates. The branch-disagreement gate gives the same held-out positive accuracy (79.94\%) while reducing ASR only modestly to 9.67\%. The surface-rule gate gives the lowest ASR among these simple deployable rules, but it pays a positive-accuracy cost, matching the LLaVA design-set warning.

\begin{table}[t]
\centering
\small
\begin{tabular}{@{}lrr@{}}
\toprule
Held-out fixed rule & ASR $\downarrow$ & Pos. acc. $\uparrow$ \\
\midrule
Safe neutralization & 0.1098 & 0.7346 \\
Surface-rule gate & \textbf{0.0464} & 0.6962 \\
Task-prior/disagree & 0.0490 & 0.7378 \\
Branch disagreement & 0.0967 & \textbf{0.7994} \\
Answer disagreement & 0.0693 & \textbf{0.7994} \\
\bottomrule
\end{tabular}
\caption{Held-out aggregate over GLM-4.6V and Llama-3.2-Vision only ($n=1572$--1576 applicable negatives for the first four rows, $n=1572$ for answer disagreement, and $n=1560$ positives). LLaVA is treated as the design model for choosing the reported fixed gates.}
\label{tab:heldout}
\end{table}

This aggregate is useful because it prevents per-model cherry-picking. If we select the best gate separately for each model, the story becomes stronger but less deployable: branch disagreement is attractive on GLM, task-prior/disagreement is attractive for Llama ASR, and the surface-rule gate minimizes ASR on both at the cost of supported commitments. A fixed-rule comparison instead asks what would happen if the same routing rule were shipped after design on LLaVA. Under that standard, task-prior/disagreement is the strongest ASR-preserving-positive rule, branch disagreement is the conservative utility-preserving alternative, and answer disagreement is the only simple category-prior-free rule that improves both aggregate axes.

The answer-disagreement router is a post-hoc candidate. It adds one check to branch disagreement: if pressure and neutral disagree but TPCD still emits a concrete commitment, the router keeps pressure. Because it uses only branch outputs, not task-category priors, it is cleaner than the task-prior rule; because it was selected after examining outputs and still relies on surface predicates, it still needs human-label validation.

The result should still be read carefully. The held-out models share the same tone-matters taxonomy, and the task-prior rule encodes category information rather than a learned visual signal. The answer-disagreement router avoids explicit category priors but is still selected post-hoc and still identifies commitment through surface forms. The improvement supports benchmark-internal mitigation rules, then, but not a universal claim that these routers will generalize to arbitrary pressure prompts. A stronger protocol would tune a continuous gate on one set of human-audited examples, freeze it, and test it on new categories and images.

\subsection{Time-Category Sensitivity}

The time subtask raises a genuine concern: time-reading positives are weak, and time categories are also where safe neutralization often fails on negatives. To address it we compute two sensitivities. First, removing positive time-reading rows from the GLM/Llama matched positives raises the aggregate positive accuracy of both task-prior/disagreement and branch-disagreement to 94.90\%, compared with 85.19\% for safe neutralization. This indicates that the utility penalty in the full positive set is concentrated in the time subtask rather than in text or object positives. Second, removing the negative time categories (4--6) lowers safe-neutralization ASR to 5.12\% and branch/task-prior ASR to 1.54\%. In this no-time negative view, the two disagreement-based gates coincide because the task prior only differs on time/intent categories; both still reduce ASR relative to neutralization.

This sensitivity changes the interpretation of the task-prior gate. Its strongest ASR gain partly comes from forcing TPCD on time categories where neutralization is unreliable, but these are also categories where current positive utility is hard to measure. At the same time, the no-time sensitivity shows that routing is not solely a clock artifact: object and text categories still benefit when pressure commits and the neutral branch withholds commitment. We therefore keep the time results in the main table, because they are part of the tone-matters threat model, but avoid claiming that the current positive controls fully establish utility for time-reading deployment.

\subsection{Qualitative Error Patterns}

Representative outputs clarify why the gates behave differently. On a Llama indeterminate-clock example, the pressure branch answers ``The time on the clock is 37:59,'' while TPCD produces ``No readable time is visible''; safe neutralization also withholds commitment. On a GLM absent-object example, pressure reasons that chalk is present because a blackboard is visible, whereas neutralization says that no chalk is directly visible and TPCD answers absent. Here pressure commits, neutral does not, and TPCD removes the schema-based inference.

The harder cases are those in which all branches are wrong in different ways. Some text examples produce strings under both pressure and neutralization, and TPCD can generate malformed repetitions rather than a clean abstention. These errors explain why full TPCD is not monotone better than safe prompting and why branch outputs are useful signals but not calibrated labels.

\section{Discussion}

\paragraph{Why TPCD differs from ICD.}
ICD uses a disturbed instruction that can alter task semantics and often strengthens yes bias on POPE. TPCD uses a pressure version of the same task; on tone-matters this branch specifically amplifies the surface commitments being attacked. This explains the aggregate gap (ICD 47.12\% ASR vs. TPCD 0.50\%), but the current evidence is still mostly behavioral. Stronger evidence would add per-token sequence logit analysis on text/time/object commitment token sets, not only first-token POPE diagnostics.

The distinction is also visible in failure modes. Generic disturbed instructions can create a negative branch unrelated to the visual question, suppressing useful task tokens or leaving pressure commitments untouched. A pressure branch keeps the task fixed and changes the demand for certainty, making it a targeted probe for tone-sensitive commitments. This targeting is imperfect: if pressure increases correct target tokens, subtraction can hurt utility, so the branch is useful only with a router for visually weak cases.

\section{Limitations}

The main LLaVA experiments use one main $\alpha$; the additional GLM/Llama full runs use the same fixed setting rather than a held-out alpha sweep. The positive controls are synthetic; the larger 780-example matched positives reduce the small-$n$ problem for cross-model checks, but time-reading accuracy remains only 50\% under pressure for GLM/Llama and 0\% in the original LLaVA positives. The text gate is operationally label-free but not independently validated, because its negative success is tied to the same surface predicates used by the scorer for several categories. POPE uses first-token yes/no scoring and a different operationalization from sequence TPCD; its JSD AUROC is based on 35 pressure-induced false-yes cases and has poor calibration. First-token JSD thresholds are exploratory and are fit on tiny diagnostic rows, not held-out deployment gates.

There are also protocol limitations in the cross-model evidence. We call GLM and Llama ``held-out'' only with respect to model family: the gate rules are fixed after LLaVA design, but the benchmark categories and scorer predicates are shared. This is weaker than holding out new tasks, new images, or human-labeled unsupportedness annotations. The task-prior/disagreement gate is especially vulnerable to this limitation because it explicitly encodes category priors; the answer-disagreement router avoids category priors but was still chosen post-hoc from the same set of generated branch outputs. The held-out aggregate should therefore be interpreted as evidence that the rules transfer across model families on the same benchmark, not as evidence that they will generalize to arbitrary pressure prompts.

Finally, the current implementation is computationally more expensive than safe prompting because it requires at least two branches and, for TPCD, stepwise logit combination. We do not optimize latency or memory, and we do not test commercial closed models where logits may be unavailable. A practical deployment would need either a cheaper proxy for pressure sensitivity or an API setting that exposes the relevant branch probabilities.

\section{Conclusion}

Pressure prompts expose a commitment distribution that TPCD can subtract, driving LLaVA tone-matters ASR from 66.75\% to 0.50\%. The harder problem is label-free routing. Surface-rule gates catch unsupported commitments on negative-only benchmarks but over-fire on supported positives; branch-disagreement gates protect positives but fall back to neutralization-level ASR on categories where neutralization also commits. Task-prior/disagreement and answer-disagreement routers show that pressure subtraction can beat neutralization on held-out model-family aggregates, and full GLM/Llama evaluation shows that this model-family transfer is not unique to LLaVA. However, the current routers remain post-hoc and surface-form based. The next step is not more optimistic framing of the current rules, but an independently validated grounding-aware router that can decide when pressure-induced commitment should actually be subtracted.

\bibliographystyle{plainnat}
\bibliography{references}

@inproceedings{rohrbach2018object,
  author       = {Anna Rohrbach and
                  Lisa Anne Hendricks and
                  Kaylee Burns and
                  Trevor Darrell and
                  Kate Saenko},
  editor       = {Ellen Riloff and
                  David Chiang and
                  Julia Hockenmaier and
                  Jun'ichi Tsujii},
  title        = {Object Hallucination in Image Captioning},
  booktitle    = {Proceedings of the 2018 Conference on Empirical Methods in Natural
                  Language Processing, Brussels, Belgium, October 31 - November 4, 2018},
  pages        = {4035--4045},
  publisher    = {Association for Computational Linguistics},
  year         = {2018},
  url          = {https://doi.org/10.18653/v1/d18-1437},
  doi          = {10.18653/V1/D18-1437},
  bibsource    = {dblp computer science bibliography, https://dblp.org}
}

@inproceedings{li2023pope,
  author       = {Yifan Li and
                  Yifan Du and
                  Kun Zhou and
                  Jinpeng Wang and
                  Wayne Xin Zhao and
                  Ji{-}Rong Wen},
  editor       = {Houda Bouamor and
                  Juan Pino and
                  Kalika Bali},
  title        = {Evaluating Object Hallucination in Large Vision-Language Models},
  booktitle    = {Proceedings of the 2023 Conference on Empirical Methods in Natural
                  Language Processing, {EMNLP} 2023, Singapore, December 6-10, 2023},
  pages        = {292--305},
  publisher    = {Association for Computational Linguistics},
  year         = {2023},
  url          = {https://doi.org/10.18653/v1/2023.emnlp-main.20},
  doi          = {10.18653/V1/2023.EMNLP-MAIN.20},
  bibsource    = {dblp computer science bibliography, https://dblp.org}
}

@inproceedings{leng2024vcd,
  author       = {Sicong Leng and
                  Hang Zhang and
                  Guanzheng Chen and
                  Xin Li and
                  Shijian Lu and
                  Chunyan Miao and
                  Lidong Bing},
  title        = {Mitigating Object Hallucinations in Large Vision-Language Models through
                  Visual Contrastive Decoding},
  booktitle    = {{IEEE/CVF} Conference on Computer Vision and Pattern Recognition,
                  {CVPR} 2024, Seattle, WA, USA, June 16-22, 2024},
  pages        = {13872--13882},
  publisher    = {{IEEE}},
  year         = {2024},
  url          = {https://doi.org/10.1109/CVPR52733.2024.01316},
  doi          = {10.1109/CVPR52733.2024.01316},
  bibsource    = {dblp computer science bibliography, https://dblp.org}
}

@article{wang2024icd,
  author       = {Xintong Wang and
                  Jingheng Pan and
                  Liang Ding and
                  Chris Biemann},
  title        = {Mitigating Hallucinations in Large Vision-Language Models with Instruction
                  Contrastive Decoding},
  journal      = {CoRR},
  volume       = {abs/2403.18715},
  year         = {2024},
  url          = {https://doi.org/10.48550/arXiv.2403.18715},
  doi          = {10.48550/ARXIV.2403.18715},
  eprinttype   = {arXiv},
  eprint       = {2403.18715},
  bibsource    = {dblp computer science bibliography, https://dblp.org}
}

@inproceedings{chuang2024dola,
  author       = {Yung{-}Sung Chuang and
                  Yujia Xie and
                  Hongyin Luo and
                  Yoon Kim and
                  James R. Glass and
                  Pengcheng He},
  title        = {DoLa: Decoding by Contrasting Layers Improves Factuality in Large
                  Language Models},
  booktitle    = {The Twelfth International Conference on Learning Representations,
                  {ICLR} 2024, Vienna, Austria, May 7-11, 2024},
  publisher    = {OpenReview.net},
  year         = {2024},
  url          = {https://openreview.net/forum?id=Th6NyL07na},
  bibsource    = {dblp computer science bibliography, https://dblp.org}
}

@inproceedings{fu2023mme,
  author       = {Chaoyou Fu and
                  Peixian Chen and
                  Yunhang Shen and
                  Yulei Qin and
                  Mengdan Zhang and
                  Xu Lin and
                  Jinrui Yang and
                  Xiawu Zheng and
                  Ke Li and
                  Xing Sun and
                  Yunsheng Wu and
                  Rongrong Ji and
                  Caifeng Shan and
                  Ran He},
  editor       = {Danielle Belgrave and
                  Cheng Zhang and
                  Laura N. Montoya and
                  Hsuan{-}Tien Lin and
                  Razvan Pascanu and
                  Piotr Koniusz and
                  Marzyeh Ghassemi and
                  Nancy Chen and
                  Iv{\'{a}}n Vladimir Meza Ru{\'{\i}}z and
                  Arturo Loaiza{-}Bonilla},
  title        = {{MME:} {A} Comprehensive Evaluation Benchmark for Multimodal Large
                  Language Models},
  booktitle    = {Advances in Neural Information Processing Systems 38: Annual Conference
                  on Neural Information Processing Systems 2025, NeurIPS 2025, San Diego,
                  CA, USA, December 2-7, 2025 / Mexico City, Mexico, November 30 - December
                  5, 2025},
  year         = {2025},
  url          = {http://papers.nips.cc/paper\_files/paper/2025/hash/d79a27cf2772fe00be7f341efc0eb517-Abstract-Datasets\_and\_Benchmarks\_Track.html},
  bibsource    = {dblp computer science bibliography, https://dblp.org}
}

@misc{amber2023,
      title={Mitigating Hallucinations in Large Vision-Language Models with Instruction Contrastive Decoding}, 
      author={Xintong Wang and Jingheng Pan and Liang Ding and Chris Biemann},
      year={2024},
      eprint={2403.18715},
      archivePrefix={arXiv},
      primaryClass={cs.CV},
      url={https://arxiv.org/abs/2403.18715}, 
}

@article{zhao2025sycophancy,
  author       = {Yunpu Zhao and
                  Rui Zhang and
                  Junbin Xiao and
                  Changxin Ke and
                  Ruibo Hou and
                  Yifan Hao and
                  Ling Li},
  title        = {Sycophancy in vision-language models: {A} systematic analysis and
                  an inference-time mitigation framework},
  journal      = {Neurocomputing},
  volume       = {659},
  pages        = {131217},
  year         = {2026},
  url          = {https://doi.org/10.1016/j.neucom.2025.131217},
  doi          = {10.1016/J.NEUCOM.2025.131217},
  bibsource    = {dblp computer science bibliography, https://dblp.org}
}

@article{tone_matters_repo,
  author       = {Zhiyuan Jiang and
                  Weihao Hong and
                  Xinlei Guan and
                  Tejaswi Dhandu and
                  Miles Q. Li and
                  Meng Xu and
                  Kuan Huang and
                  Umamaheswara Rao Tida and
                  Bingyu Shen and
                  Daehan Kwak and
                  Boyang Li},
  title        = {LLM-as-Judge Framework for Evaluating Tone-Induced Hallucination in
                  Vision-Language Models},
  journal      = {CoRR},
  volume       = {abs/2604.18803},
  year         = {2026},
  url          = {https://doi.org/10.48550/arXiv.2604.18803},
  doi          = {10.48550/ARXIV.2604.18803},
  eprinttype   = {arXiv},
  eprint       = {2604.18803},
  bibsource    = {dblp computer science bibliography, https://dblp.org}
}

\end{document}